\documentclass{article}

\usepackage[utf8]{inputenc} 
\usepackage[T1]{fontenc}    
\usepackage{hyperref}       
\usepackage{url}            
\usepackage{booktabs}       
\usepackage{amsfonts}       
\usepackage{nicefrac}       
\usepackage{microtype}      
\usepackage{xcolor}         
\usepackage{thmtools}
\usepackage{amsmath}
\usepackage{tikz}
\usepackage{listings}
\usepackage[capitalise]{cleveref}

\declaretheorem{definition}
\declaretheorem{example}

\title{Measuring global properties of neural generative model outputs via generating mathematical objects}

\author{
  Bernt Ivar Utstøl Nødland \\
  Norwegian Defence Research Establishment (FFI)\\
  Instituttveien 20,  2007 Kjeller \\
  Norway \\
  \texttt{bernt-ivar-utstol.nodland@ffi.no} \\
 }

\begin{document}

\maketitle

\begin{abstract}
We train deep generative models on datasets of reflexive polytopes. This enables us to compare how well the models have picked up on various global properties of generated samples. Our datasets are complete in the sense that every single example, up to changes of coordinate, is included in the dataset. Using this property we also perform tests checking to what extent the models are merely memorizing the data. We also train models on the same dataset represented in two different ways, enabling us to measure which  form is easiest to learn from. We use these experiments to show that deep generative models can learn to generate geometric objects with non-trivial global properties, and that the models learn some underlying properties of the objects rather than simply memorizing the data.
\end{abstract}

\section{Introduction}

In recent years techniques for generating artifical data has improved, particularly by using deep learning techniques.
Well-known examples of this are generative adversarial networks \cite{GAN}, variational auto-encoders \cite{VAE} and generators based on the transformer \cite{attention}. 
Evaluating the performance of most such models is a difficult problem, since it is not clear which metrics would accurately determine that such a generative model is performing well. A salient property of outputs of these models, which partly motivates this paper, is that  frequently generated data  locally looks like plausible data, while there is some global mistake that reveals that the generated data is not really plausible. Examples are generated images of people with 3 front teeth, or generated text where the pronoun of a person changes throughout the text. We here propose to train generative models on datasets of mathematical objects, all satisfying some global properties. This enables us to measure whether the generative model has picked up on the global properties, or if it simply generates locally plausible data.

The datasets we consider in this paper already exists in the academical literature: In the intersection of mathematics and theoretical physics there has been considerable interest in reflexive polytopes, which has led to the construction of  databases of reflexive polytopes \cite{kreuzerskarke},\cite{polydb},\cite{obro}: A database containing a complete list of all smooth reflexive lattice polytopes of dimensions up to nine, as well as a database of all reflexive lattice polytopes of dimensions up to four. These datasets are large enough that they can be used to train high performing deep learning models. They  are also very well suited to our purposes of checking global properties, since all polytopes of a given dimension share several non-trivial global properties. 

These datasets are   complete in the sense that
 they include every single example (up to changes of coordinate) satisfying the required properties, thus ensuring that the training data is in fact accurately representative of the data in question. This also enables us to study to what extent the models are memorizing the training data or are learning some underlying structure of the data: For each generated sample satisfying the required properties we can check whether the data defining the sample is identical to the corresponding example in the training data, or if it is a coordinate change of it.
 
There are two equivalent ways of giving the data of a polytope: Either as the set of points  satisfying some linear inequalities or as the convex hull of some vertices. This property enables us to train generative models on these different representations of the data and check from which representation the model  learns most.

We believe that training neural networks on datasets of mathematical objects is an interesting way to test hypotheses about generative models, because one can check the mathematical properties of network outputs. The members of our datasets are discrete objects, thus one might hope that a model's performance on them will be analogous to its performance on other discrete sources of data, like text. A limitation of this approach is that there may be important ways in which mathematical datasets are different from real world data such as text. On the other hand, the preciseness of mathematics gives some advantages over real world data: Since the datasets are complete, the training data will be completely representative of the actual data, and since the two equivalent ways of specifying a polytope denote exactly the same object, we can check which representation is easiest to learn from without worrying about impreciseness in translating between them.

This paper is in addition a contribution to the exploration of how one could use techniques from machine learning to problems in pure mathematics: We train  generative models on datasets of geometric objects and study to what extend the models can generate examples with similar properties. We speculate that neural generative models might in the future be used to generate counterexamples to mathematical conjectures, for instance in cases where generating random examples is not straight-forward.

\section{Related work}

There is a large body of literature about how to evaluate generative models. In particular for text models, which, by the discrete nature of their data, is quite similar to that of discrete mathematical datasets, there are several different metrics one can use to measures performance, see the survey \cite{evaluationsurvey} for details. This paper is not strictly speaking about this problem, however it is partly motivated by the fact that measuring performance of generative models is difficult, and suggests ways to measure how well different model architectures learn global properties.

The paper \cite{mathreason3} is of a similar flavour to ours in that it suggests using mathematical datasets to measure the performance of neural architectures. However, they focus on whether a model can answer correctly mathematical questions like arithmetic or solving equations.
In a similar direction, there are recently  several other papers attempting to use deep learning to solve mathematics questions. Some examples include the paper \cite{symbolicmath}, which studies how one can use deep learning models to solve differential equations, as well as how to solve symbolic integration problems and the paper \cite{lamplec}, which studies how to predict properties of differential equations using neural networks. 

Szegedy \cite{Szegedy} argues that using techniques from natural language understanding is an essential step towards creating systems of autoformalization: computer programs that can prove mathematical theorems. 
This is an ambitious program, well outside the scope of this paper, however our paper is a small step in the exploration of how techniques from natural language processing could be used in computer programs that handles mathematical objects. 

In a different direction, this paper can also be considered a continuation of already existing efforts to explore how to apply deep learning techniques to datasets in the intersection of geometry and physics. Several papers have already tried to use classification techniques for mathematical datasets, including datasets of Calabi-Yau manifolds (these are closely linked to reflexive polytopes): This was initiated in \cite{He1} and has been followed up in several papers, see for instance the paper \cite{Hesurvey} and the included references. These efforts have so far been mostly towards exploring to what extent deep learning models can correctly predict properties of geometric objects from some defining data.

\section{Datasets}

\subsection{Motivation for using mathematical datasets} \label{secion:motivation}

A hard problem in the study of generative models is the question of how one should measure performance of the model.
We suggest that an interesting way of measuring the performance of generative model architectures would be to test on datasets of mathematical objects: If one trains on a dataset of objects all satisfying a property P, such that syntactically similar objects could very well fail to satisfy property P, one can test to what extent a generative model manages to pick up on the  property P. It is not clear to what extent differences in performance in picking up on properties of mathematical objects correspond to similar differences in performance on other types of data (say text or images); this would be an interesting future research question.

We are primarily interested in global properties. By the term "global property" we heuristically mean a  property that is a property of the whole object, and not a property that is localized in a specific place. More mathematically rigorous we are thinking of properties that are not local-to-global: A property of an object is said to be local-to-global if it is the case that for some subdivision into local pieces, if every local piece has the property then the global piece also has the property. In language an example of a local-to-global property is that of being grammatically correct; if each sentence in a text is grammatically correct then the entire text is grammatically correct. On the other hand, narrative coherence is not a local-to-global property, since one easily can string together texts that are each narratively coherent into a text that is not. When we say global property in this paper we will mean a property that is either not local-to-global, or for which the local property does not even make sense.

Another reason why using datasets of mathematical objects could be useful is motivated by the following frequent challenge for deep learning models: One trains a model on some dataset that is supposed to be representative of real world data, however, in practice it is virtually impossible to collect such a representative dataset. For this reason it is hard to tell which part of a model's mistakes come from the fact that the dataset does not accurately portray the underlying data generating mechanism and what comes from the model's inability to learn the structure of the data. For some mathematical datasets (including the ones we use in this paper) this problem can be avoided: If the class of mathematical objects satisfying a property P is finite, then one could train the dataset on the complete list (or some random fraction) of examples. Then no mistake the model makes can  be put down to bias in the distribution of the training  data.

\subsection{Details on datasets of polytopes}

In \cref{appendix:polytope} we give mathematical background on polytopes and their properties. We essentially suggest two different datasets of polytopes that can be used to train deep generative models: The Kreuzer-Skarke database of relexive polytopes of dimension $4$ \cite{kreuzerskarke} and the database of smooth reflexive polytopes of dimension up to $9$ \cite{polydb}. In the latter case the dimensions $8$ and $9$ are probably the most reasonable to train large deep models on, since there in those cases are a large number of examples.

For the experiments in this paper we choose as our dataset all smooth reflexive convex lattice polytopes of dimension $7$ and of dimension $8$. This is because these datasets are large enough to train good models, but small enough that training do not take a lot of time on a single computer. The 7d dataset contains $72256$ samples, while the 8d dataset contains $749 892$ samples. Each coordinate entry is regarded as a token, thus $"0"$ is a possible token (in fact the most common) but also "-2" is one token. One could have chosen alternative tokenizations, for instance by considering each inequality as a token or each individual character a token. The vocabulary size will be around $30$ with our chosen convention. The max sequence length of the 7d dataset is 169 tokens and 219 tokens for the 8d dataset. If one would like to scale up the experiments of this paper we consider both the $9$-dimensional smooth reflexive polytopes and the  $4$-dimensional reflexive polytopes to be interesting datasets, since these are significantly larger than our chosen datasets.

A data sample is a matrix, where each row corresponds to $w \in \mathbb{Z}^d$ corresponding to the inequality  $1+w \cdot x \geq 0$. Together all these inequalities define a polytope.  Each polytope $P$ in the dataset satisfy the global properties of being a lattice polytope, compact, reflexive, smooth and normal.  Since the dataset is complete, it is also true that any generated example satisfying all these properties  will be in the training dataset, up to change of coordinates. In this case we call the generated example "correct".

The properties cannot be seen independently of each other: If $P$ is not a lattice polytope we automatically have that $P$ is not smooth. If $P$ is a lattice polytope defined by inequalities of the form $1+w_i \geq 0$, then it will automatically be reflexive. If $P$ is reflexive and smooth, then it is automatically normal (for $\dim P \leq 8$). If $P$ is not compact, then normality is not defined, so we consider it as not normal. If $P$ is smooth, then by conjecture $P$ is always normal, although this is not proven in general. For generated polytopes it is then interesting to check for the following properties:

\begin{itemize}
    \item $P$ is compact
    \item $P$ is a lattice polytope (hence also reflexive)
    \item $P$ is smooth
    \item $P$ is normal
\end{itemize}

Above we assume that the polytope is defined via inequalities. However, there is also a different way of giving the data of a convex lattice polytope: By giving the vertices and stating that the polytope is the convex hull of  these. We also train generative models on a subset of the 7d dataset where we give the polytope via vertices. By construction all samples will then automatically be convex lattice polytopes. The properties we test for under this formulation is therefore:

\begin{itemize}
    \item $P$ is reflexive
    \item $P$ is smooth
    \item $P$ is normal
\end{itemize}

Training on these two different data representations can be considered  analogous to stating the same semantic content in different languages.
\subsection{Sensitivity to local changes}

\begin{figure}
\noindent\begin{minipage}{.49\textwidth}
\begin{lstlisting}
 -1 0 0 0 0 0 0 0
 0 0 0 0 0 0 0 -1
 0 -1 0 0 0 0 0 0
 0 0 0 0 -1 0 0 0
 0 0 0 0 -1 1 0 1
 0 0 0 0 0 -1 0 0
 0 0 -1 0 0 0 0 0
 0 0 0 -1 0 0 0 0
 0 0 0 0 0 0 -1 0
 1 1 1 1 0 -4 1 1
 0 0 0 0 1 0 <@\textcolor{red}{0}@> -1
\end{lstlisting}
\end{minipage}\hfill
\noindent\begin{minipage}{.49\textwidth}
\begin{lstlisting} 
 -1 0 0 0 0 0 0 0
 0 0 0 0 0 0 0 -1
 0 -1 0 0 0 0 0 0
 0 0 0 0 -1 0 0 0
 0 0 0 0 -1 1 0 1
 0 0 0 0 0 -1 0 0
 0 0 -1 0 0 0 0 0
 0 0 0 -1 0 0 0 0
 0 0 0 0 0 0 -1 0
 1 1 1 1 0 -4 1 1
 0 0 0 0 1 0 <@\textcolor{red}{1}@> -1
\end{lstlisting} 
\end{minipage}

\caption{Each example in the dataset is a matrix of integers such as the one on the left. The matrix on the right is not valid, since we changed one entry and it turns out the new matrix does not satisfy the global properties.}
\label{figure:datasample}
\end{figure}

The data sample on the left in \cref{figure:datasample} is taken from the dataset. Thus it defines a normal smooth  reflexive lattice polytope. The data on the right is the same as on  the left, except one zero in the last row has been changed to a one. It still defines a compact polytope, but it no longer has any of the properties reflexive, smooth, normal, nor being a lattice polytope. This illustrates that global properties of the data is highly non-trivial to deduce simply from the local structure. 

To give an indication of how volatile the properties are, we drew 1000 random examples from the dataset and did a random change of either changing a single $0$ to $\pm 1$ or changing a single $\pm 1$ to a $0$. The distribution of the properties in the modified polytopes were:
\begin{itemize}
\item Compact: 851 
\item Lattice polytope: 789 
\item Smooth: 227 
\item Normal: 642 
\item All: 147
\end{itemize}
Thus only 147 out of 1000 are still correct datasamples with a single small change. With more than one change even less polytopes will be correct. This indicates that a successful model has to understand global properties, since the modified samples are locally plausible examples of the dataset.

\section{Experiments} \label{section:experiments}

All models are trained with one polytope being one data sample. We add special start-of-sequence and end-of-sequence tokens to each sample and pad to the fixed maximum length. 
We train until the loss is reasonably stable, which is somewhere in the range 100-800 epochs, depending on size of model and dataset. We train with the Adam optimizer with learning rate $0.001$. The models are trained on a single Titan V GPU with approximately 110 hours of combined training time. Checking the global properties for one list of 1000 8d polytopes took about 6 hours using a single CPU.  We do not do an extensive search of hyperparameters, since the experiments are more intended as a proof of concept than as a claim about the best possible performance. Thus we could probably train models that were even better performing by increasing model sizes and using other  network architectures and traning procedures. We have chosen to keep things as vanilla as possible, not least because we really only want a good performing model: Since the size of the vocabulary is small and the training datasets not extremely big one risks that a very high performing model simply memorizes all the samples rendering the outputs uninteresting. Thus a good model giving mostly reasonable outputs is what we wish to study. This is also the most reasonable analogy with other generative models, since in most situations training a generative model on the complete list of possible examples is not possible.

\subsection{Generating polytopes} \label{subsection:genpoly}

Our most fundamental experiments are training  generative models on the 8d dataset. After training the models we generate 1000  polytopes using each model and use the open source computer algebra software systems Macaulay2 \cite{M2} and polymake  \cite{polymake:2000}, \cite{polymake:2017} to check all the global properties in question. For the generated polytopes satisfying all of the  properties (which implies that they are, up to change of coordinates,  in the training set) we check whether they are exact copies of their representative in the training set. We also check whether they are the same as their representative up to reordering of rows. The defining inequalities is in reality a set and not a sequence, thus reordering the rows is the simplest form of coordinate change. The models are trained on the data samples as sequences, thus there are no a priori reason the model should "understand" that the rows are really sets. 

\subsection{Baseline}
To compare the results with a baseline we also do the following. We make a simple model that generates samples based on sampling from the histogram of tokens following the previous $n$ tokens, where $n$ is some natural number (we choose $n=10$ in our experiments). Thus this is a model that simply generates the next token based on the frequency of tokens preceding the given token in the training data. In particular this is by construction a local model.  Therefore it cannot use the complete history to generate the next token, hence it cannot know the total length of the sequence, which leads to artificially short samples. We remedy this by replacing the newline token to also include the line number, thus the model cannot generate the end of sequence token before at least $8d+1$ steps (since there are no such samples in the dataset). This fix will lead to samples that have a chance of satisfying the properties and that looks locally plausible.

\subsection{Training on half the dataset}

We also train models on half of the 7d and 8d datasets (chosen randomly) and generate polytopes from the models. Then we can check whether the polytopes generated by these models are to a large extent from the half it was trained on, or whether they are equally from both halves. These tests give some indication of to what extent the models are simply memorizing the dataset, or whether they learn some underlying structures.

\subsection{Training on convex hull representation}

We also train models on the convex hull representation of the dataset and compare the performance to that of the original hyperplane   representation. For these experiments we use the 7d dataset, since the samples are longer in the convex hull representation, and thus  require more computational resources both to train the models and to check the properties of generated samples. In the hyperplane representation, a reflexive  polytope of dimension $d$ can have at most $3d$ defining inequalities \cite[Theorem 3]{casagrande}. For the number of vertices there are no such simple upper bound
, and in practice they are indeed much longer, the longest example in the 7d dataset having more than 3000 tokens. We choose the subset of the 7d dataset consisting of those examples that have less than 800 tokens in the convex hull representation, to avoid very long sequences. This dataset has 39737 examples, thus it is approximately half the 7d dataset. We train generative models on both this and the corresponding hyperplane dataset and compare performance on the two datasets.

\subsection{Implementation}

We train recurrent networks with one or two LSTM layers. We use an embedding dimension of 4 
and hidden dimension 256 for each of the LSTM layers. 

We also train transformer models that are stacked encoder layers. We choose an embedding dimension of 128 and use 4 multi-attention heads and stack 4 such encoder layers to get the full model. 

The models are trained to predict the next token in the dataset. We generate examples by sampling from the output probabilities until we generate the end-of-sequence token.
The sizes of the models are chosen so that the transformer and the 2-layer LSTM models will have approximately the same number of parameters.

\section{Results}

\subsection{Training on the full 8d dataset}

In \cref{table:8dbase}  we see the how many of the generated samples satisfied the different global properties. In \cref{table:8dpermutation} we see how many of the correctly generated samples  that equalled their representative in the training set precisely and up to permutation of rows. In \cref{table:8dintersection} we see how many generated examples satisfied some choices of several properties at once.

\begin{table} [hbt!] 
  \begin{tabular}{lllll}

 \bf{Model} & \bf{LSTM 1} & \bf{ LSTM 2} & \bf{Transformer} & \bf{Baseline} \\ 
    \hline
    All properties & 238 & 260 & 404 & 2 \\
        Compact & 827 & 840 & 909 & 21 \\
            Lattice polyhedron & 712 & 722  & 832 & 412 \\
                    Smooth & 270 & 304 & 440 & 178 \\
                      Normal  & 590  & 613 & 755 & 2\\
                          \hline

  \end{tabular}
  \caption{For each model we see how many of the 1000 generated samples that satisfy the various properties, including the condition all of them at once.} \label{table:8dbase}
\end{table}

\begin{table} [hbt!] 
  \begin{tabular}{lllll}
    
 \bf{Model} & \bf{LSTM 1} & \bf{ LSTM 2} & \bf{Transformer} & \bf{Baseline} \\
    \hline
    Copy  & 0/238  & 0/260  & 0/404 & 0/2 \\
            Permutation  & 36/238 = 15.1 \%  & 58/260=22.3 \%  & 77/404 = 19.1 \% & 0/2 \\
        \hline
  \end{tabular}

  \caption{Any example satisfying all properties exist in the training dataset, up to  change of coordinates. Here we see how many of the generated samples are exact copies of their representative in the training set and how many are copies up to permutation of rows.} \label{table:8dpermutation}
\end{table}

\begin{table} [hbt!] 
  \begin{tabular}{lllll}
    
 \bf{Model} & \bf{ LSTM 1} & \bf{LSTM 2} & \bf{Transformer} & \bf{Baseline}\\

    \hline
    Compact+lattice  & 590 & 613 & 755 &2 \\
                Compact+lattice+normal & 590 & 613  & 755 & 2 \\
                      Compact+not smooth  & 589  & 580 & 505 & 19 \\
                    Compact+lattice+not smooth & 352 & 353& 351 & 0 \\

                          \hline

  \end{tabular}

  \caption{We here see a breakdown of some of the intersections of the properties. In other words, either how many examples that satisfy several of the properties, or their negations. Note the equality of  the two first rows are purely empirical, since mathematically there are no reason why they should be equal because there exist (lots of) examples of reflexive lattice polytopes that are not normal}\label{table:8dintersection}
\end{table}

\subsection{Training on half of the 8d dataset}

For the models trained on half the 8d dataset the results for the properties of generated samples is  seen in \cref{table:halfproperties} and  intersections in \cref{table:8dhalfintersection}.

\begin{table} [hbt!] 
  \begin{tabular}{llll}

 \bf{Model} & \bf{1-layer LSTM} & \bf{2-layer LSTM} & \bf{Transformer} \\ 
    \hline
    All properties & 187 & 244 & 361 \\
        Compact & 783 & 834 & 908  \\
            Lattice polyhedron & 667 & 706  & 851 \\
                    Smooth & 237 & 282 & 396 \\ 
                      Normal  & 522  & 590 & 772 \\
                          \hline

  \end{tabular}

  \caption{For each model trained on half the dataset we see how many of the 1000 generated samples that satisfy the various properties, including the condition all of them at once.} \label{table:halfproperties}
\end{table}

\begin{table} [hbt!] 
  \begin{tabular}{llll}
    
 \bf{Model} & \bf{1-layer LSTM} & \bf{2-layer LSTM} & \bf{Transformer} \\
    \hline
    Compact+lattice  & 522 & 590 & 772 \\
                Compact+lattice+normal & 522 & 590  & 772 \\
                      Compact+not smooth  & 596  & 590 & 547 \\
                      Compact+lattice+not smooth & 335 & 346 & 411 \\

                          \hline

  \end{tabular}

  \caption{We here see a breakdown of some of the intersections of the properties for the half 8d dataset.}
   \label{table:8dhalfintersection}
\end{table}

The primary motivation for also training models on half the data is that it enables us to measure to which half of the dataset generated samples belong. For a given correctly generated sample  we know that an equivalent representation of the polytope has to be in the training data. However, checking which of the approximately 800 000 examples it is equivalent to is computationally expensive: Doing so for one example takes approximately 1 day on the computer these experiments were run on, thus we have not been able to do this at large scale. However, we can quickly check the examples that are equal to their original representative up to permutation of rows, seen in \cref{table:halffraction}. To get more substantial numbers we also train a transformer on half the 7d dataset and generate 1000 polytopes, of which 319 are correct. For all of these we manage to check which training sample a generated sample is equivalent to (checking one example took about 1 hour)  and find that 179/319 (approximately 56 \%) of generated examples are from the half  the model was trained on.

\begin{table} [hbt!] 

 \begin{tabular}{llll}

 \bf{Model} & \bf{LSTM 1} & \bf{LSTM 2} & \bf{Transformer} \\
    \hline
    In half 8d dataset  & 15/28 = 53.6 \%   & 16/37 = 43.2 \%  & 42/73=57.5 \% \\
        \hline

  \end{tabular}
  \caption{Among the generated samples that are equal to a permutation of the representation in the full dataset, this table shows the fraction of samples that were in the half of the dataset the model was actually trained on.} \label{table:halffraction}
\end{table}

\subsection{Convex hull representation}

We train Transformer and 2-layer LSTM on the subset of 7d polytopes that are less than 800 tokens in the convex hull representation and on the hyperplane representation of the same subset. In \cref{table:conv} we see the results for the convex hull representation, while \cref{table:convreference} shows the corresponding results for the hyperplane representation.

\begin{table} [hbt!] 
  \begin{tabular}{lll}

 \bf{Model}  & \bf{2-layer LSTM} & \bf{Transformer} \\ 
    \hline
    All properties & 7 & 285 \\
                          Reflexive & 7  & 285 \\
                             Smooth  & 7 & 286 \\
                                 Normal    & 27 & 410 \\
                          \hline

  \end{tabular}

  \caption{For each model trained on the 7d convex hull dataset we see how many of the 1000 generated samples that satisfy the various properties, including the condition all of them at once.} \label{table:conv}
\end{table}

\begin{table} [hbt!] 
  \begin{tabular}{lll}
      
 \bf{Model}  & \bf{2-layer LSTM} & \bf{Transformer} \\ 
    \hline
    All properties & 216 & 547 \\
        Compact& 660 & 909  \\
             Lattice polyhedron & 616  & 850 \\
                      Smooth  & 288 & 583 \\
                       Normal    & 421 & 784 \\
                          \hline

  \end{tabular}

  \caption{For each model trained on the subset of 7d hyperplane dataset that is short in the convex hull representation we see how many of the 1000 generated samples that satisfy the various properties, including the condition all of them at once.} \label{table:convreference}
\end{table}

\section{Discussion} \label{section:discussion} 

We see that the models successfully manage to generate many examples with all of the correct properties. This shows that the models manage to pick up on global properties. The naive baseline using only local properties performs significantly worse than the neural models with almost no correct samples. We also see that the transformer models generally perform better than the LSTM models.  Unsurprisingly some properties are shown to be much easier to understand than others, with smoothness being the hardest. We also observe that the models trained on half the 8d dataset generate almost the same (or even higher) number of compact, normal and lattice polyhedra as the models trained on the full dataset, but less smooth polytopes. Thus it seems that doubling the amount of data improves performance on harder properties, but not on easier properties.
Several of the properties we study are   expensive to actually compute: Our best known methods to computationally check that a polytope is a lattice polytope  are highly non-trivial \cite[Chapter 1]{ziegler}. To check that a polytope is smooth one would additionally compute the determinant of the edges emanating from each vertex. Thus it is  unlikely that the models actually manage to perform operations  approximating these computations. Rather, it seems  more likely that the network are relying on vague heuristic observations that correlate with the properties. For instance, for a polyhedron to be compact it is necessary that all variables of the defining inequalities are bounded. In particular every variable has to appear with both positive and negative coefficient in the inequalities defining the polyhedron. This translates into the property that every column has at least one positive and one negative entry. Thus, this is a necessary property of all the samples, however it is not sufficient for the polyhedron being compact. It seems reasonable to suspect that the model relies on this kind of heuristic necessary patterns that correlates with the property in question, also for the more complicated properties. Clearly it is possible to formulate similar heuristics for other properties and, maybe, given enough such patterns, one manages to reliably generate examples with correct properties without actually verifying the  properties.

Somewhat surprisingly the models trained on the hyperplane representation generated no examples of compact lattice polytopes that were not normal. This might indicate that there could be some underlying mathematical reason why lattice polytopes that are similar (in the vague sense of being generated by our model) to smooth compact reflexive lattice polytopes are always normal.  
We wonder whether understanding why we get no non-normal lattice polytopes might reveal some underlying structure, that could potentially help in proving that all smooth (reflexive) polytopes are normal? Empirical observations such as this indicates that using generative models could be useful in the field of experimental mathematics: Unexpected properties of generated objects might give one hints about new statements that one could subsequently try to prove.

We also observe that almost none of the generated samples are  identical to their representative in the dataset. This gives  evidence that the model is not simply memorizing the data, but rather understanding some underlying structure. However, a substantial portion (roughly 1 in 5) of the samples are equal to their representative up to permutation of rows. This proportion is way higher than one would expect from a random permutation. This could be an indication that the model does perform some amount of memorization. However, the examples in the training data are generated via the algorithm of \cite{obro} and  by using this there are systematic patterns in how the polytopes are represented in the training data. Hence it is not very clear that a random permutation is a reasonable comparison, and thus we are not able to draw very clear  conclusions  from this.

The experiments using half the data, while not having enough data to be very certain,  indicate that the models are actually "understanding" some underlying structure, rather than simply memorizing the training data: The fraction of correctly generated samples that are in the training data are only slightly exceeding 50 \% (56 \% for the largest set of examples we managed to check). If the model did a large amount of memorization one would expect to see a large majority of samples from the half the models were trained on.

When given the dataset in the convex hull representation the models pick up on significantly less patterns than they do in the hyperplane representation. The LSTM model in particular almost did not generate any examples with the correct properties. An obvious reason for this is that the sequences are longer. However, these models also had some factors in their favour, since the space of possible objects are smaller.
From these experiments one might suspect that detecting properties is harder in the convex hull representation.

\section{Future work}

There are several instances where we think it would be interesting to try generative models on mathematical datasets. An obvious example is to use other types of generative models such as generative adversarial networks  and variational autoencoders. Another obvious extension is using generative models on other matematical datasets, for instance on interesting continuous  datasets. A third idea is to investigate to what extent  encoding group invariance in the model improves performance: if one builds into the model that the rows are sets and not sequences, or more ambitiously that there is a  matrix group of coordinate changes that the samples are invariant under, one would expect performance to improve.

One of the motivations for working on this project, which we were not able to complete, was to construct a counterexample to Oda's conjecture in convex geometry/algebraic geometry: Does there exist a smooth lattice polytope that is not normal? \cite{oda} The reason we believed this might be worth a try is that generating random examples of smooth polytopes in not easy \cite{bruns}. This is also a challenge for those who wishes to use generative models to generate new examples, since there does not exist a large training set of arbitrary smooth polytopes. 
We believe a reasonable future attempt would be to try transfer learning: Pre-train a generative model to generate non-normal lattice polytopes and then fine-tune with smooth polytopes to maybe discover a counterexample. Of course, it might be the case that Oda's conjecture is true and no such example exists. Nonetheless, we suspect that these techniques could in the future be used for finding counterexamples in mathematics.
The paper \cite{graphcounter}  is an example of this, where counterexamples in graph theory are discovered using neural generative models trained using reinforcement learning.

The behaviour of global properties of mathematical objects under the pre-train then fine-tune methodology could also be of independent interest: If one pre-trains generative models on mathematical objects satisfying property P and fine-tunes on similar objects satisfying Q, where most examples that satisfy Q does not satisfy P, does the model tend to generate objects satisfying both P and Q or does it tend to generate examples satisfying Q and not P?

\section{Conclusion}

We have shown that neural generative models successfully can generate  discrete mathematical objects with interesting global properties. We have presented evidence that indicates that the models actually pick up on some of the underlying global structures. We believe that these  techniques could be  useful in both the evaluation of  generative models and in the field of experimental mathematics. 

\bibliography{ref}
\bibliographystyle{alpha}






\appendix


\section{Mathematical background on polytopes} \label{appendix:polytope}

Here we give a short introduction to the mathematics of polytopes and their properties, required to understand the datasets we use. 

\begin{definition}
A polyhedron is a subset of $\mathbb{R}^d$ defined by a finite number of linear inequalities. In other words a set of the form  
\[ \{ x \in \mathbb{R}^d | b+ Ax \geq 0 \}, \]
for some matrix $A$ and vector $b$. A polyhedron is called a lattice polyhedron if all vertices of $P$ are integer points.
\end{definition}

The dimension of a polyhedron is defined as the dimension of the smallest linear subspace containing the polyhedron. A polyhedron is called convex if it forms a convex subset of $\mathbb{R}^d$.

\begin{definition}
A polyhedron which is compact (meaning it has finite volume) is called a polytope.
\end{definition}

In this paper we will study polyhedra up to lattice equivalence. In other words we consider two polyhedra $P$ and $Q$ as isomorphic if they are equal after some affine coordinate change of $\mathbb{R}^d$ that equals a translation term plus an element of $GL_d(\mathbb{Z})$. In other words a sum of an integer translation and a change of basis matrix with integer entries having determinant equal to $\pm 1$. 

Convex lattice polytopes has received a lot of attention in the intersection of toric geometry and string theory. To explain this connection we need to define the following property of a polytope:

\begin{definition}
To a lattice polyhedron $P$ one can associate the dual polyhedron $P^\vee$ defined by:
\[ P^\vee = \{ u \in \mathbb{R}^d | x \cdot u  \geq -1 \text{ for all } x \in P \}, \]
where $\cdot$ is the ordinary scalar product of vectors. A lattice polyhedron is called reflexive if the associated dual polyhedron is also a lattice polyhedron. 

\end{definition}

For a lattice polytope the condition that $P^\vee$ is also a lattice polytope is equivalent to the condition $P$ has a unique interior integer point. To a reflexive polytope one can associate a toric Fano  variety. Inside these toric Fano varieties one can construct so-called Calabi-Yau manifolds. According to superstring theory the universe is a 10-dimensional manifold, where four dimensions are space and time dimensions, while the remaining six dimensions corresponds to a three-dimensional complex Calabi-Yau manifold (three complex dimensions corresponds to six real dimensions). The conjectural relationship between a Calabi-Yau and the dual Calabi-Yau associated to the dual polytope is what is known as \emph{mirror symmetry}.
Because almost all known examples of three-dimensional Calabi-Yau manifolds live inside four-dimensional toric fano manifolds,  string theorists have been very interested in studying four-dimensional reflexive polytopes
This is the background for the construction of the Kreuzer-Skarke database of all reflexive polytopes of dimensions $3$ and $4$. Parallel to this development there has also been significant interest in reflexive polytopes in the mathematics community.

\begin{definition}
A lattice polyhedron is called smooth if for any vertex the minimal integer points of the emanating edges form a $\mathbb{Z}$-basis for the integer points of $\mathbb{R}^d$. In other words, there has to be exactly $d$ edges emanating from each vertex and the matrix containing the minimal integer vectors of these edges  has determinant $\pm 1$.
This condition is equivalent to the condition that the toric variety associated to the polyhedron is a smooth manifold.
\end{definition}

 Of a given dimension there are only finitely many  reflexive polytopes, up to integer change of coordinates. Øbro developed an algorithm to find all smooth reflexive polytopes of a given dimension \cite{obro} which was used to construct the database of all smooth lattice polytopes up to dimension $9$. An additional property a lattice polytope can have is that of normality:

\begin{definition}
A lattice polytope $P$ is called normal  if it has the property that all integer lattice points of any integer multiple $lP$ is the sum of $l$ integer lattice points from $P$.
\end{definition}

It is a famous open conjecture that states that any smooth polytope is normal \cite{oda}. For reflexive polytopes this has been checked up to dimension $8$; thus all polytopes of dimension less than $8$ in the database of smooth relexive polytopes are normal.

\begin{example}
We here show an example of a $2$-dimensional polytope. It can be described as the set of points satisfying the inequalities below. It is compact since it has finite area. It is a lattice polytope since all its vertices are integer points. We remark that the inequalities being defined by inequalities with integer coefficients are far from being sufficient to imply that the vertices are integer points. It is reflexive since the only interior integer point is the origin. It is normal since all $2$-dimensional lattice polytopes are normal \cite[Proposition 1.2.4]{normal}. We store the data defining such a polytope in terms of the inequalities defining it, where we omit the constant $1$, since the constant term always can be assumed to equal $1$ by \cite[Theorem 8.3.4]{CLS}. Equivalently we could also store it by recording the vertices: In that case we recover the polytope by taking the convex hull of the vertices.
\end{example}
\noindent\begin{minipage}{.24\textwidth}
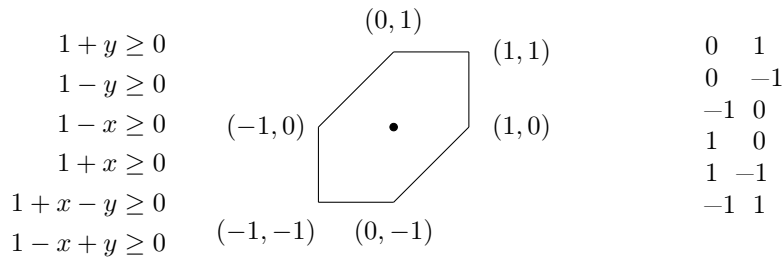
\begin{align*}
    1+y &\geq 0 \\
    1-y &\geq 0 \\
    1-x &\geq 0 \\
    1+x &\geq 0 \\
    1+x-y &\geq 0 \\
    1-x+y &\geq 0 
\end{align*}
\end{minipage}
\noindent\begin{minipage}{.5\textwidth}
\begin{tikzpicture} [scale=1.0]
\node[draw,circle,inner sep=1pt,fill] at (0,0) {};
\draw [-] (1,0) -- (1,1);
\draw [-] (1,1) -- (0,1);
\draw [-] (0,1) -- (-1,0);
\draw [-] (-1,0) -- (-1,-1);
\draw [-] (-1,-1) -- (0,-1);
\draw [-] (0,-1) -- (1,0);
\node at (1.7,0) [] {$(1,0)$};
\node at (1.7,1) [] {$(1,1)$};
\node at (0,1.4) [] {$(0,1)$};
\node at (-1.7,0) [] {$(-1,0)$};
\node at (-1.7,-1.4) [] {$(-1,-1)$};
\node at (0,-1.4) [] {$(0,-1)$};
\end{tikzpicture}
\end{minipage}
\noindent\begin{minipage}{.24\textwidth}
\begin{lstlisting}
  0  1
  0  -1
  -1 0
  1  0
  1 -1
  -1 1
\end{lstlisting}
\end{minipage}
Left: defining inequalities. Center: polytope drawn with vertices. Right: how we store inequalities.

\section{Examples}

Here we give some examples of data samples. Below we see two examples from the 7d dataset, both in the hyperplane 
representation and in the convex hull representation.

\begin{figure}[hbt!]
\noindent\begin{minipage}{.49\textwidth}
\begin{lstlisting}
 0 0 0 0 0 0 1 
 0 0 0 0 0 0 -1 
 -1 0 0 0 0 0 0 
 0 -1 0 0 0 0 0 
 0 0 -1 0 0 0 0 
 0 0 0 -1 0 0 0 
 0 0 0 0 -1 0 0 
 0 0 0 0 0 -1 0 
 1 1 1 1 1 1 6 
\end{lstlisting}
An example from the 7d dataset in the hyperplane representation.
\end{minipage}\hfill
\noindent\begin{minipage}{.49\textwidth}
\begin{lstlisting} 
 1 1 1 1 1 1 1 
 -12 1 1 1 1 1 1 
 1 -12 1 1 1 1 1 
 1 1 -12 1 1 1 1 
 1 1 1 -12 1 1 1 
 1 1 1 1 -12 1 1 
 1 1 1 1 1 -12 1 
 1 1 1 1 1 0 -1 
 1 1 1 1 0 1 -1 
 1 1 1 0 1 1 -1 
 1 1 0 1 1 1 -1 
 1 0 1 1 1 1 -1 
 0 1 1 1 1 1 -1 
 1 1 1 1 1 1 -1
\end{lstlisting} 
The same example in the convex hull representation.
\end{minipage}
\end{figure}

\begin{figure}[hbt!]
\noindent\begin{minipage}{.49\textwidth}
\begin{lstlisting}
 0 0 0 0 -1 1 0 
 0 0 0 0 -1 0 0 
 0 -1 0 0 0 0 0 
 -1 0 0 0 0 0 0 
 0 0 -1 0 0 0 0 
 0 0 0 -1 0 0 0 
 0 0 0 0 0 0 -1 
 0 0 0 0 0 -1 0 
 0 0 0 0 0 1 0 
 1 1 1 1 1 -6 1 
\end{lstlisting}
An example from the 7d dataset in the hyperplane representation.
\end{minipage}\hfill
\noindent\begin{minipage}{.49\textwidth}
\begin{lstlisting} 

 1 1 1 1 1 1 0 
 1 1 1 1 1 0 1 
 -6 1 1 1 1 0 1 
 1 -6 1 1 1 0 1 
 1 1 -6 1 1 0 1 
 1 1 1 -6 1 0 1 
 1 1 1 1 0 1 1 
 1 1 1 0 1 1 1 
 1 1 0 1 1 1 1 
 1 0 1 1 1 1 1 
 0 1 1 1 1 1 1 
 1 1 1 1 1 1 1 
 1 1 1 1 0 -1 1 
 -11 1 1 1 0 -1 1 
 1 -11 1 1 0 -1 1 
 1 1 -11 1 0 -1 1 
 1 1 1 -11 0 -1 1 
 1 1 1 1 -12 -1 1 
 1 1 1 1 1 0 -6 
 1 1 1 1 0 -1 -11 
\end{lstlisting} 
The same example in the convex hull representation.
\end{minipage}
\end{figure}

\section{Remarks on semantically ambiguous datasets}

Relevant to the discussion in \cref{section:discussion}  is the slightly philosophical observation that the data in and of itself does not carry the semantic meaning of the inequalities. The model might "interpret" it as some other structure and generate similar examples to these other structures. In our case there is actually quite obviously two different intuitive ways of interpreting the data: For a reflexive polytope, the duality between a polytope $P$ and its dual $P^\vee$ can be seen explicitly as follows: If  $P$ is given by hyperplane inequalities $1+w_i \cdot x \geq 0$, then $P^\vee$ is the polytope that is the convex hull of the $w_i$. Strictly speaking $P^\vee$ is living in the dual vector space to where $P$ is living, but the model has no way to know this. Mathematically, some of the properties of $P$ might be checked by verifying a corresponding property for the dual $P^\vee$. Thus even if the model were to actually do mathematical operations on the actual geometric object, it is not clear which geometric object it would do them on. This even shows that the distinction between the hyperplane representation and the convex hull representation does not really exist: Every hyperplane sample also represents the dual polytope in the convex hull representation and, dually, every sample in the convex hull representation also represents the dual in the hyperplane version.

A consequence of this is that the very question of what our training dataset really is, is problematized: We have claimed that our dataset represents all smooth reflexive lattice polytopes of a given dimension in the hyperplane representation. It could equally well represent some subset of all reflexive polytopes of the given dimension in the convex hull representation. Note that these datasets are distinct, since the dual of a smooth reflexive polytope is not necessarily smooth. Which one of these interpretations the model is "understanding" is not clear, or if it is either a combination of the two or even some third semantic interpretation. Maybe this shows that the notion of the model "understanding" the semantics of the syntactic representation of the data is simply ill-founded. On this view the model only ever sees numbers and syntactically manipulates them according to correlations, and all questions of semantic interpretations are put on top by humans.

\end{document}